\documentclass{pprai}

\usepackage[utf8]{inputenc}
\usepackage[T1]{fontenc}
\usepackage{graphicx}
\usepackage{amsthm}
\usepackage{txfonts}
\usepackage{url}
\usepackage{algpseudocode}

\title{Face Consistency Benchmark for GenAI Video}
\headtitle{Face Consistency Benchmark for GenAI Video}

\author{
Michal Podstawski$^{1[0000-0003-1222-6894]}$\\
Malgorzata Kudelska$^{1[0009-0008-7560-5238]}$\\
Haohong Wang$^{2[0009-0007-7829-9232]}$
}

\headauthor{M. Podstawski, M. Kudelska, H. Wang}
\affiliation{%
  $^1$TCL Research Europe\\
  Grzybowska 5A, 00-132 Warsaw, Poland\\
  $^2$TCL Research America\\
  2025 Gateway Pl, San Jose, CA, USA\\
\{name.surname\}@tcl.com}


\begin{document}
\maketitle

\begingroup
\renewcommand\thefootnote{}\footnote{\hspace{-1.8em}Presented at PP-RAI 2025: 6th Polish Conference on Artificial Intelligence, April 7–9, 2025, Katowice, Poland.}
\addtocounter{footnote}{-1}
\endgroup

\begin{abstract}
    Video generation driven by artificial intelligence has advanced significantly, enabling the creation of dynamic and realistic content. However, maintaining character consistency across video sequences remains a major challenge, with current models struggling to ensure coherence in appearance and attributes. This paper introduces the Face Consistency Benchmark (FCB), a framework for evaluating and comparing the consistency of characters in AI-generated videos. By providing standardized metrics, the benchmark highlights gaps in existing solutions and promotes the development of more reliable approaches. This work represents a crucial step toward improving character consistency in AI video generation technologies.
\end{abstract}

\section{Introduction}

The rapid advancement of artificial intelligence (AI) has profoundly transformed video generation, enabling the creation of realistic and dynamic scenes with minimal human input. These innovations have had a major impact on industries such as entertainment, advertising, and education, providing powerful tools for creativity and automation. As a result, AI-generated videos now exhibit increasingly complex environments, natural movements, and improved scene composition, pushing the boundaries of what synthetic media can achieve.

Despite these achievements, one critical challenge remains unresolved: the consistent generation of characters across video sequences. Current AI models often struggle to maintain coherence in character appearance and attributes when generating videos, leading to visual inconsistencies that detract from the overall quality and usability of the content. These inconsistencies hinder the adoption of AI-generated video technologies in applications that require precise storytelling, character-driven narratives, or high-quality animation.

To address this limitation, this paper introduces the Face Consistency Benchmark (FCB), an evaluation framework designed to measure and compare the ability of AI models to generate consistent facial representations of characters. The benchmark provides standardized evaluation metrics, enabling researchers and developers to objectively assess existing solutions and identify key areas for improvement.

\section{Related work}

Recent advancements in AI-generated video have led to the development of various benchmarks to evaluate the quality and performance of video generation models. These benchmarks provide standardized methods for assessing aspects such as realism, temporal coherence, and visual fidelity, enabling researchers to compare and improve generative models effectively.

One example is AIGCBench~\cite{fan2024aigcbenchcomprehensiveevaluationimagetovideo}, a comprehensive benchmark designed to evaluate the capabilities of state-of-the-art video generation algorithms. It provides a diverse, open-domain image-text dataset that allows for the assessment of various algorithms under standardized conditions. AIGCBench employs 11 metrics across four key dimensions - control-video alignment, motion effects, temporal consistency, and video quality - offering a robust evaluation framework. These metrics include both reference-dependent and reference-free evaluations, ensuring a thorough and versatile analysis of algorithm performance.

Another notable example is VBench~\cite{huang2023vbenchcomprehensivebenchmarksuite}, an extensive benchmark suite designed to evaluate video generative models. VBench decomposes video generation quality into 16 well-defined dimensions, including subject identity inconsistency, motion smoothness, temporal flickering, and spatial relationships, facilitating fine-grained and objective evaluation. For each dimension, VBench provides tailored prompts and evaluation methods, ensuring a thorough assessment of model performance. Additionally, VBench includes human preference annotations to validate the alignment of its benchmarks with human perception, offering valuable insights into the strengths and weaknesses of current video generation models.

However, while existing benchmarks such as AIGCBench and VBench offer comprehensive evaluation frameworks for video generation, their focus primarily lies on aspects like motion quality, temporal consistency, and overall video realism. They do not specifically address character facial consistency, a crucial element for achieving realism in character-driven videos. This gap highlights the need for specialized benchmarks that emphasize facial consistency, fostering more robust advancements in AI-generated video content.

\section{Proposed solution}

To address the challenge of character facial consistency in AI-generated videos, this paper proposes a dedicated evaluation framework, the Face Consistency Benchmark (FCB). Unlike existing benchmarks, FCB is specifically designed to measure the ability of video generation models to maintain consistent facial features. By focusing on face similarity metrics, FCB provides a robust tool for assessing how well models preserve identity, expressions, and fine details, which are crucial for achieving realism in character-driven content. This targeted approach bridges a critical gap in the evaluation of AI video generation and facilitates meaningful advancements in the field.

Proposed framework achieves its goal by utilizing commonly used face recognition models, including VGG-Face~\cite{Parkhi2015DeepFR}, Facenet, Facenet512~\cite{Schroff_2015}, ArcFace~\cite{Deng_2022}, SFace~\cite{boutros2022sfaceprivacyfriendlyaccurateface}, and GhostFaceNet~\cite{ghostface}. To seamlessly integrate and handle these models, the framework leverages DeepFace library~\cite{serengil2024lightface}. Selected models are well-suited for evaluating facial similarity and consistency, as they are designed to extract robust features representing identity and expressions. By leveraging these state-of-the-art models, the proposed benchmark ensures accurate and reliable assessments of character facial consistency in AI-generated videos, enabling a thorough comparison of video generation models.

The paper evaluates four text-to-video generation models. Three of them are open-source: HunyuanVideo~\cite{kong2024hunyuanvideo}, Vchitect-2.0~\cite{vchitect}, and CogVideoX1.5-5B~\cite{yang2024cogvideox}. The other model, Runway Gen-3~\cite{runway}, is accessible through APIs. These models were selected as they ranked among the top performers on the VBench benchmark at the time of writing, ensuring the analysis highlights the most advanced video generation systems available.

For each model, 30 videos were generated using a consistent set of prompts derived from real videos to ensure a fair comparison. The prompts were created with the help of ChatGPT~\cite{chatgpt}, utilizing frames from the real videos. The base videos were specifically chosen to represent a diverse range of subjects, including variations in gender, age, and lighting conditions, which were reflected in the generated prompts. Additionally, they featured movements that challenge generative AI, such as head rotations and vertical motions, ensuring that the evaluation effectively tests the ability to handle complex motion dynamics.

To standardize evaluation across all models, the maximum resolution was set to 720×720, as each model can generate videos at this resolution or higher. For models that produced videos with higher resolutions, the outputs were adjusted accordingly to ensure consistency in evaluation. Importantly, the evaluation focused on the face, with cropping performed from the entire frame. If a face was not detectable in a frame (e.g., when the character was turned away), that frame was skipped to maintain relevance in the assessment.

\begin{figure}[ht]
\centering
\includegraphics[width=1\textwidth]{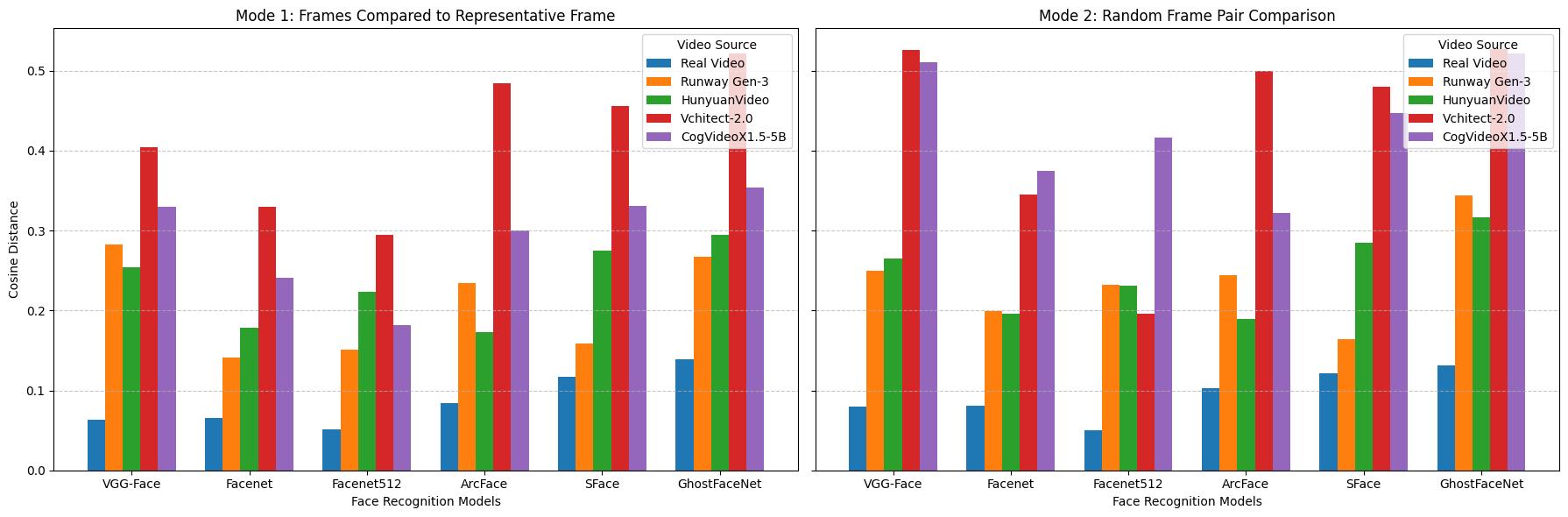}
\caption{Comparison of face consistency in real and AI-generated videos. The evaluation verifies video generation models using similarity between the face in different frames, measured by cosine distance (lower is better). \textit{Mode 1} (left) compares all frames to a representative frame. \textit{Mode 2} (right) assesses temporal consistency through random frame pairs. Results are averaged over 30 videos.}
\label{fig:cosine_distances}
\end{figure}

The evaluation consists of two modes of comparison. In the first mode, all frames from a video are compared to a selected representative frame, which serves as the reference model for the character's face (Table \ref{tab}). This approach focuses on assessing the similarity of generated frames to the expected character face. In the second mode, 200 random pairs of frames are compared within each video, with individual frames potentially repeating across pairs (Table \ref{tab2}). This method evaluates the coherence of character faces across frames, ensuring consistency throughout the entire video. Both modes use the cosine distance of facial embeddings as the metric, where lower values indicate greater similarity (if appropriate, it can be easily switched to Euclidean or L2-normalized Euclidean distance). To provide a baseline for comparison, we also measure real videos using the same methodology, allowing us to better contextualize the performance of AI-generated videos. Together, these modes provide a comprehensive assessment of both facial accuracy and temporal consistency in generated videos. Results are shown in Figure \ref{fig:cosine_distances}.

\begin{table}[!ht]
\caption{Cosine distance of facial embeddings for the first mode of comparison, where all frames are compared to a selected representative frame.}
\label{tab}
\centering
\footnotesize 
\setlength{\tabcolsep}{5pt} 
\renewcommand{\arraystretch}{1.3} 
\begin{tabular}{|l|c|c|c|c|c|c|}
\hline
\textbf{Source} & \textbf{VGG-Face} & \textbf{Facenet} & \textbf{Facenet512} & \textbf{ArcFace} & \textbf{SFace} & \textbf{GhostFaceNet} \\ 
\hline\hline
Real Video & 0.0636 & 0.0650 & 0.0514 & 0.0843 & 0.1267 & 0.1391 \\ 
\hline
Runway Gen-3 & 0.2827 & \textbf{0.1408} & \textbf{0.1511} & 0.2346 & \textbf{0.1584} & \textbf{0.2668} \\ 
HunyuanVideo & \textbf{0.2542} & 0.1784 & 0.2229 & \textbf{0.1734} & 0.2746 & 0.2946 \\ 
Vchitect-2.0 & 0.4042 & 0.3295 & 0.2951 & 0.4843 & 0.4554 & 0.5215 \\ 
CogVideoX1.5-5B & 0.3294 & 0.2412 & 0.1813 & 0.3005 & 0.3310 & 0.3541 \\ 
\hline
\end{tabular}
\end{table}

\begin{table}[!ht]
\caption{Cosine distance of facial embeddings for the second mode of comparison, where 200 random frame pairs are compared within each video.}
\label{tab2}
\centering
\footnotesize 
\setlength{\tabcolsep}{5pt} 
\renewcommand{\arraystretch}{1.3} 
\begin{tabular}{|l|c|c|c|c|c|c|}
\hline
\textbf{Source} & \textbf{VGG-Face} & \textbf{Facenet} & \textbf{Facenet512} & \textbf{ArcFace} & \textbf{SFace} & \textbf{GhostFaceNet} \\ 
\hline\hline
Real Video & 0.0798 & 0.0805 & 0.0498 & 0.1027 & 0.1119 & 0.1308 \\ 
\hline
Runway Gen-3 & \textbf{0.2493} & 0.1987 & 0.2319 & 0.2441 & \textbf{0.1641} & 0.3441 \\ 
HunyuanVideo & 0.2655 & \textbf{0.1955} & 0.2307 & \textbf{0.1896} & 0.2842 & \textbf{0.3161} \\ 
Vchitect-2.0 & 0.5255 & 0.3447 & \textbf{0.1962} & 0.4997 & 0.4798 & 0.5266 \\ 
CogVideoX1.5-5B & 0.5101 & 0.3744 & 0.4162 & 0.3215 & 0.4469 & 0.5213 \\ 
\hline
\end{tabular}
\end{table}

This experiment underscores the persistent challenges in achieving character facial consistency in AI-generated videos. While HunyuanVideo and Runway Gen-3 showed relatively better performance compared to other models, they still fall significantly short of real video consistency. Their lower cosine distances indicate some ability to maintain similarity and coherence, yet the gap remains substantial. These findings highlight the limitations of current generative video models and emphasize the need for further research to improve character realism and temporal consistency.

\section{Next steps}

Future work could enhance evaluation in two key directions. First, extending benchmarks to multi-character settings would allow for the detection and assessment of individual characters in complex scenes, addressing challenges like interactions and occlusions. Second, broadening evaluation to include full-body coherence - encompassing posture, limb movement, and overall character dynamics - would provide a more holistic measure of realism. These directions would deepen insights and foster advancements in AI video generation.

\section{Conclusions}

This paper addresses the challenge of maintaining character facial consistency in AI-generated videos by introducing the Face Consistency Benchmark (FCB). Unlike existing benchmarks, FCB focuses specifically on evaluating facial similarity and coherence across video sequences using widely adopted face recognition models.

\section*{Acknowledgment}

This manuscript acknowledges the use of ChatGPT~\cite{chatgpt}, powered by the GPT-4o language model developed by OpenAI, to improve language clarity, refine sentence structure, and enhance overall writing precision.

\bibliography{pprai}
\bibliographystyle{pprai}

\end{document}